\documentclass[runningheads]{llncs}
\usepackage[T1]{fontenc}
\usepackage{graphicx}
\usepackage{threeparttable}
\usepackage{multirow}
\usepackage{makecell}  
\usepackage{array}
\newcolumntype{C}[1]{>{\centering\arraybackslash}p{#1}}
\usepackage{amsmath}
\usepackage{amssymb}
\usepackage{caption}  
\usepackage{subcaption} 
\usepackage[table]{xcolor}
\usepackage{booktabs}       
\usepackage{hyperref}
\usepackage{color}

\begin{document}
\title{SBS: Enhancing Parameter-Efficiency of \\Neural Representations for Neural Networks \\via Spectral Bias Suppression}
\titlerunning{SBS}
%
\author{Qihu Xie \and 
Yuan Li \and
Yi Kang\thanks{Corresponding author}}
\authorrunning{Q. Xie et al.}
%
\institute{University of Science and Technology of China\\
\email{\{xieqihu,ly549826\}@mail.ustc.edu.cn}, \email{ykang@ustc.edu.cn}}
\maketitle              
\begin{abstract}
Implicit neural representations have recently been extended to represent convolutional neural network weights via neural representation for neural networks, offering promising parameter compression benefits. However, standard multi-layer perceptrons used in neural representation for neural networks exhibit a pronounced spectral bias, hampering their ability to reconstruct high-frequency details effectively. In this paper, we propose SBS, a parameter-efficient enhancement to neural representation for neural networks that suppresses spectral bias using two techniques: (1) a unidirectional ordering-based smoothing that improves kernel smoothness in the output space, and (2) unidirectional ordering-based smoothing aware random fourier features that adaptively modulate the frequency bandwidth of input encodings based on layer-wise parameter count. Extensive  evaluations on various ResNet models with datasets CIFAR-10, CIFAR-100, and ImageNet, demonstrate that SBS achieves significantly better reconstruction accuracy with less parameters compared to SOTA.

\end{abstract}
\section{Introduction}

The universal approximation theorem \cite{ref10,ref11} endows neural networks with remarkable representation capabilities that allow them to effectively fit various types of signals, including 3D scenes \cite{ref1,ref2,ref3}, images \cite{ref4}, and videos \cite{ref5}. Recently, Ashkenazi et al. introduced the concept of Neural Representation for Neural Networks (NeRN) \cite{ref6}, extending implicit neural representation (INR) to the generation of convolutional neural network (CNN) kernels. Specifically, this approach leverages the implicit representation ability of multi-layer perceptron (MLP) and incorporates knowledge distillation to train a five-layer coordinate-based ReLU-MLP. This MLP takes the spatial coordinates corresponding to a set of (layer, filter, channel) parameters within a pre-trained CNN as input and outputs the corresponding kernel.

NeRN offers a compact and continuous representation of neural network weights, introducing a new paradigm for network storage and compression. This method is particularly valuable for the mobile deployment of neural networks and provides novel insights for the design of specialized accelerators. For example, C-Transformer \cite{ref7} employs a NeRN-based approach to reduce external memory access in hardware, thus significantly improving the energy efficiency of the accelerators.

Currently, NeRN still exhibit significant room for improvement in terms of parameter efficiency. For example, to ensure that the accuracy drop of the reconstructed model on the ImageNet dataset remains within 1\%, the MLP used for training needs to have nearly half the number of parameters of ResNet-18. \cite{ref8} proposed an enhanced knowledge distillation method to improve the performance of NeRN-reconstructed models. It decouples distillation from the original training and employs a stronger teacher network to guide MLP weight generation. While conceptually sound, its practical value is limited—if a powerful teacher is available, directly fine-tuning the original model on the target data is simpler and often more effective. Additionally, the method’s reliance on task-specific, high-capacity teachers limits its generalizability.


To achieve a general improvement in NeRN's parameter efficiency without relying on external resources,  it is essential to investigate and leverage the intrinsic properties of its internal structure. Therefore, to enhance the parameter efficiency of INRs in NeRN, where coordinate-based ReLU-MLPs CNN kernels, we propose systematic architectural refinements to address spectral limitations in both output and input spaces. Our methodology builds upon the intrinsic spectral bias of standard MLPs \cite{ref9}, which act as low-pass filters preferentially encoding low-frequency patterns while attenuating high-frequency components.

\begin{figure}[h]               
  \centering                       
  \includegraphics[width=0.98\textwidth]{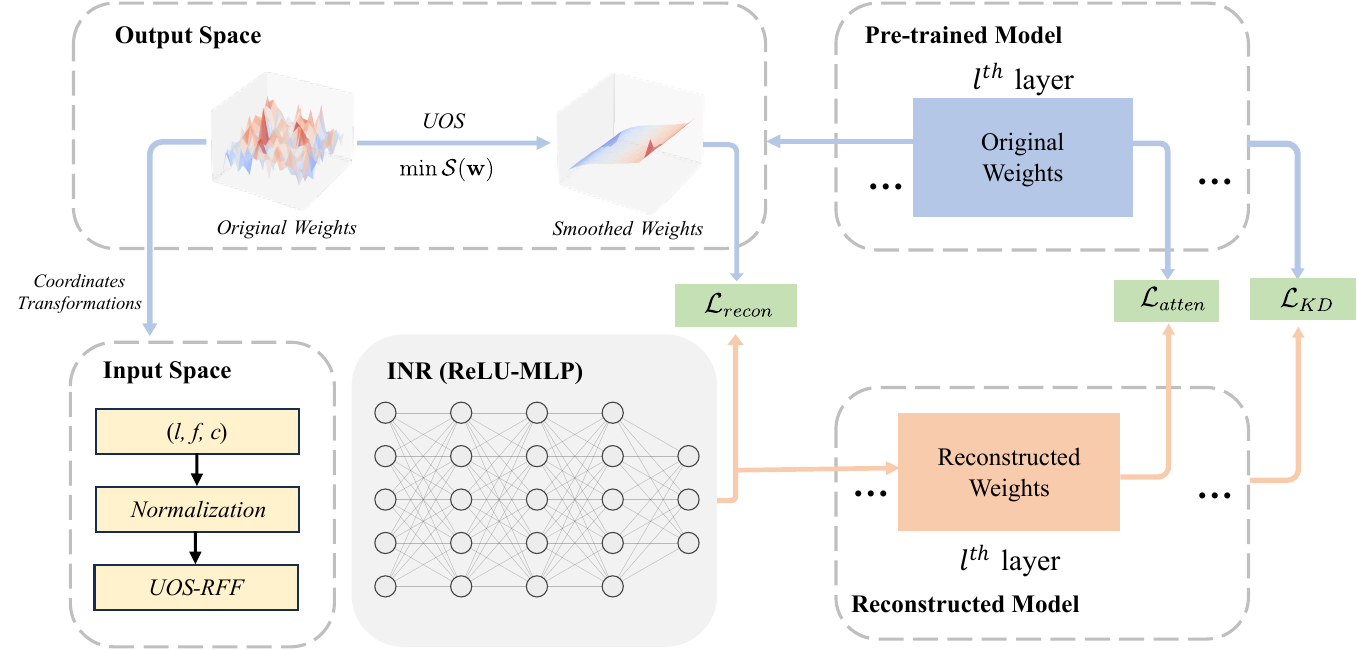}
  \caption{The framework of SBS. A single MLP is trained as an INR to generate CNN layer weights. $\mathcal{L}_\text{recon}$ is the MSE between generated and smoothed weights, $\mathcal{L}_\text{atten}$ is the MSE of feature maps across layers, and $\mathcal{L}_\text{KD}$ is the distillation loss between model outputs. The overall training objective is defined as $\mathcal{L}_\text{recon} + \alpha\mathcal{L}_\text{atten} + \beta\mathcal{L}_\text{KD}$.} 
  \label{fig:overflow}               
\end{figure}

Specifically, in the output space, the metric $\mathcal{S}(\mathbf{w})$ is defined to quantify the smoothness of the CNN kernels, which serves as the criterion for evaluating smoothness enhancement. Based on the analysis that shows that unidirectional smoothing achieves better performance compared to multi-directional smoothing, we propose a Unidirectional Ordering-based Smoothing (UOS) strategy, aiming to maximize low-frequency energy in the output space to suppress spectral bias. 


In the input space, coordinate encoding is argument with controlled spectral diversity through Random Fourier Feature (RFF). An analysis in this article on bandwidth ($\sigma^2$) reveals an inverse relationship between optimal $\sigma^2$ values and network parameter counts, then based on the analysis, we introduce the UOS-aware Random Fourier Features variant, named UOS-RFF, where the parameter $\sigma^2$ is reduced proportionally to the layer-wise parameter count in order to rebalance the frequency spectrum under UOS.

In summary, our contributions are as followings: (1) We analyze that a standard MLP behaves as a low-pass filter based on neural tangent kernel (NTK) theory, providing a theoretical foundation for NeRN; (2) we introduce structural optimizations in both the input and output spaces, to effectively alleviate spectral bias and enhance parameter efficiency; and (3) we propose the Spectral Bias Suppression framework for NeRN (SBS), as shown in Fig.~\ref{fig:overflow}, which aims to improve reconstruction performance by enhancing the capacity of the MLP in NeRN to represent more complex signals. The experimental results confirm that SBS achieves better parameter efficiency than SOTA.

\section{Related Work}



\textbf{Spectral Bias}. The universal approximation theorem \cite{ref10,ref11} establishes that neural networks with nonlinear activation functions can approximate any continuous function. However, studies have shown that neural networks exhibit a spectral bias, favoring the learning of low-frequency components over high-frequency ones \cite{ref9}. Analyzing the training dynamics, \cite{ref12,ref13} demonstrates that two-layer MLPs learn different frequency components at varying rates. To mitigate this bias, \cite{ref3,ref14} introduces positional encoding to enhance the representation of high-frequency signals. Furthermore, \cite{ref15} empirically confirms that, in image regression tasks, Random Fourier Feature (RFF) outperforms positional encoding, provided that the sampling variance is appropriately chosen. \cite{ref18,ref26} demonstrates that employing sine activation functions enables neural networks to effectively model high-frequency details.

\noindent\textbf{Implicit Neural Representations} (INRs) have emerged as a powerful paradigm for modeling continuous signals across various domains. Their compactness and ability to capture fine-grained details have led to widespread adoption in numerous applications. In the realm of image processing, INRs have been utilized for tasks such as super-resolution \cite{ref16,ref17}, inpainting \cite{ref18}, zero-shot denoising \cite{ref19}, compression \cite{ref4}, and interpolation \cite{ref20}. Beyond static images, they have been applied to video encoding \cite{ref5} and camera pose estimation \cite{ref21}. In the context of 3D shape representation, INRs have been used to map 3D coordinates to occupancy values \cite{ref22,ref23,ref24}, effectively capturing the geometry of objects. An alternative approach characterizes shapes using Signed Distance Functions (SDFs) \cite{ref25}, which provide a continuous representation of surfaces. Notably, INRs have also been employed to represent the CNN kernels \cite{ref6}, showcasing their versatility. 

\noindent\textbf{Weight generation} involves using a additional predictor network to generate the weights of a neural network. Ha et al. proposed using a smaller network, called a HyperNetwork \cite{ref27}, to predict the weights of a larger network. The HyperNetwork is trained to perform the task while learning input vectors for weight generation. \cite{ref28} extended this idea by studying the balance between prediction accuracy and diversity. In contrast, \cite{ref6} directly represents a pre-trained neural network using fixed inputs and introduces Neural Representation for Neural Networks (NeRN), which builds upon implicit neural representations (INRs). \cite{ref8} introduces an additional teacher network to improve the parameter efficiency of NeRN. In contrast, our work enhances the parameter efficiency through theoretically guided structural adjustments to the input and output spaces, without requiring any additional supervision. Beyond INRs, other network architectures have been explored for weight generation. \cite{ref29} and \cite{ref30} utilize graph neural networks (GNNs) to predict the weights of previously unseen architectures by modeling them as graph inputs. More recently, diffusion models have also been applied to model weight prediction \cite{ref31}.

\section{MLPs in NeRN}

\begin{figure}[h]               
  \centering                       
  \includegraphics[width=0.99\textwidth]{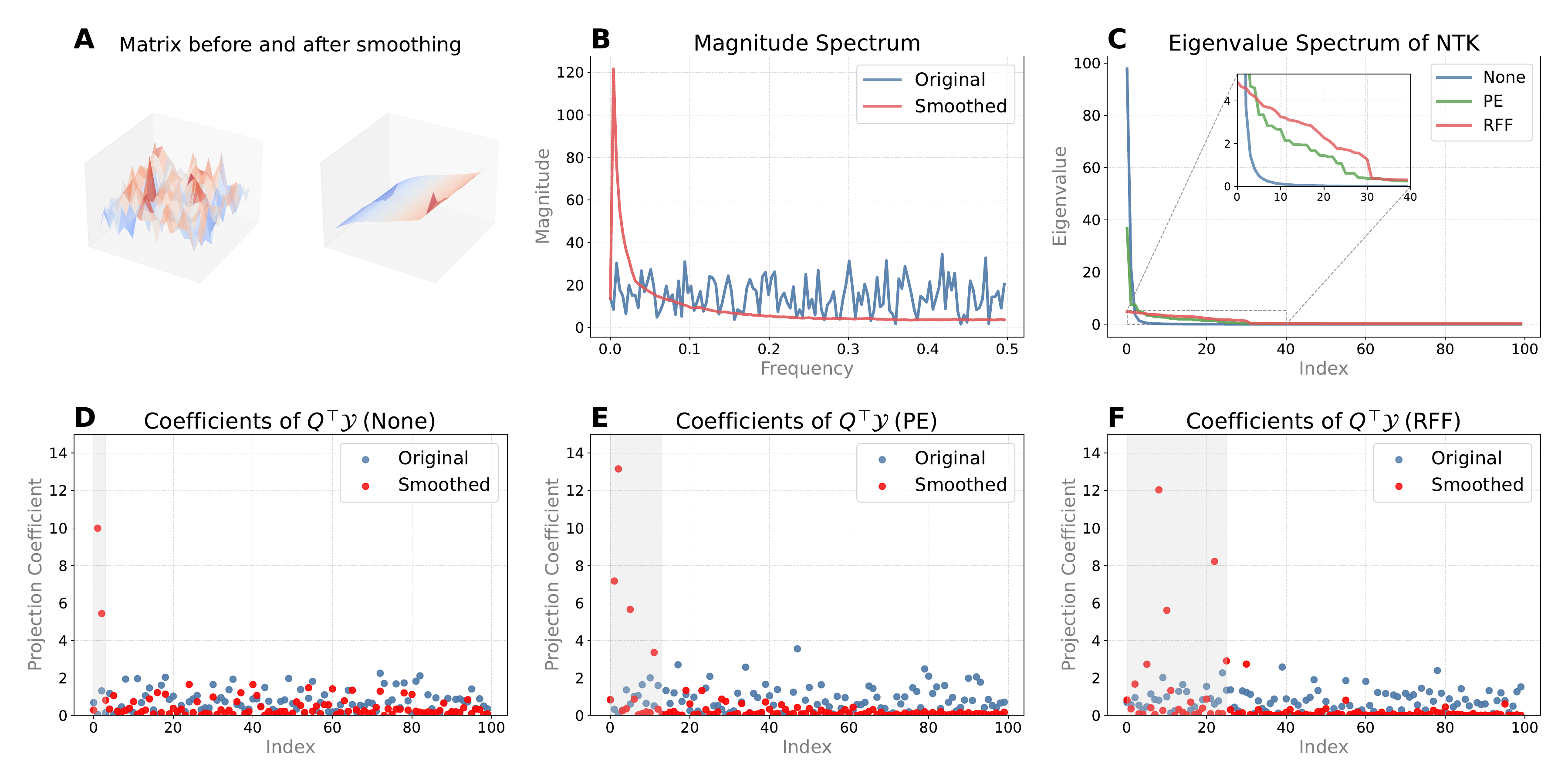}
  \caption{The impact of various coordinate transformations on the regression speed of ReLU-MLPs, for both smooth and non-smooth functions, is analyzed within the framework of NTK theory. (A) Original (left) and order-based smoothed (right) \(16\times16\) Gaussian random matrices. 
(B) Magnitude spectra obtained by FFT of the original and smoothed matrices, illustrating the concentration of energy in lower frequencies after smoothing.  
(C) Eigenvalue spectra of \(H\) under three coordinate transformations: with no encoding, \(H\) exhibits high eigenvalues only at low frequencies followed by exponential decay at higher frequencies; both PE and RFF attenuate this decay, with RFF yielding the most pronounced effect.  
(D–F) Projection coefficients of the original and smoothed \(\mathcal{Y}\) onto the eigenvectors of \(H\) across frequencies for no encoding, PE, and RFF. Smoothing increases the low-frequency coefficients, and the effective bandwidth grows in the order None < PE < RFF, indicating that RFF enables the MLP to capture richer high-frequency components.} 
  \label{fig:sample1}               
\end{figure}

In this section, the low-pass filtering properties of a standard MLP with ReLU activation are briefly analyzed based on the theory of Neural Tangent Kernel (NTK). This MLP serves as the backbone of the INRs employed in NeRN.

A ReLU-activated MLP \( f_\theta: \mathbb{R}^d \rightarrow \mathbb{R} \), trained over a dataset \( \mathcal{D} = \{(\mathbf{x}_i, y_i)\}_{i=1}^n \), is considered. Since parameter updates induce only small changes in output during training, the model can be analyzed using a linearized approximation. For MSE loss, the following expression is written:
\begin{equation}
f_t(\mathcal{X}) = \left( I - e^{-\eta H t} \right) \mathcal{Y} + e^{-\eta H t} f_0(\mathcal{X})
\end{equation}
Here, \(f_t(\mathcal{X})\) denotes the output of the MLP at time \(t\), while \(\mathcal{X}\) and \(\mathcal{Y}\) denote the inputs and the labels. The \(\eta\) represents the learning rate, and \( H(\mathcal{X}, \mathcal{X}) = \nabla_\theta f_t(\mathcal{X}) \nabla_\theta f_t(\mathcal{X})^\top \) represents the neural tangent kernel (NTK) matrix, which is a \(n \times n\) Gram matrix. The element \( H_{ij} = \langle \nabla_\theta f(\mathbf{x}_i), \nabla_\theta f(\mathbf{x}_j) \rangle \) captures the alignment of gradients with respect to each data point.

Given the symmetry of \(H\), we can apply eigendecomposition \( H = Q \Lambda Q^\top \), where \( Q \) is an orthonormal matrix and \(\Lambda\) is a diagonal matrix of non-negative eigenvalues. Without loss of generality, it is assumed that \(f_0(\mathcal{X}) = 0\), from which it follows that
\begin{equation}
Q^\top \left( f_t(\mathcal{X}) - \mathcal{Y} \right) \approx e^{-\eta \Lambda t} Q^\top \mathcal{Y}
\label{eigendecomposition}
\end{equation}
According to Previous studies \cite{ref12,ref13}, for ReLU-MLPs in the infinite-width limit and under the NTK regime, the kernel entries can be expressed as:
\begin{equation}
H(\mathbf{x}_i, \mathbf{x}_j) \approx \frac{1}{2\pi} \cos \theta_{ij} \left( \pi - \theta_{ij} \right)
\end{equation}
Here, \(\theta_{ij}\) denotes the angle between \(\mathbf{x}_i\) and \(\mathbf{x}_j\), and \(H\) represents a rotation-invariant kernel. According to Mercer's theorem, the eigenfunctions of \(H\) are the spherical harmonics \(Y_{\ell m}(x)\), with eigenvalues \(\lambda_{\ell}\) that depend only on the degree \(\ell\) and are independent of \(m\). Due to the smoothness of the ReLU kernel, the higher frequency components (i.e. larger \( \lambda_{\ell} \)) decay rapidly, as shown in Fig.~\ref{fig:sample1}~(C). Thus, the neural network tends to favor learning low-frequency components early in training, while high-frequency details are suppressed. The signal at time \( t \) can be decomposed as:
\begin{equation}
f_t(\mathcal{X}) = \sum_{\lambda_i\gg0} c_i \left( 1 - e^{-\eta \lambda_i t} \right) \mathbf{q_i}
\label{Low-Pass}
\end{equation}
Here, \( \mathbf{q_i} \) denotes the basis component in the \(i\)-th frequency direction, and \( c_i \) the projection coefficient of the target function in that direction. Since \( \lambda_i \) decays exponentially, higher frequencies contribute much less to early predictions and the MLP acts as a \textbf{Low-Pass Filter}.

\section{Spectral Bias Suppression}

Owing to the low-pass properties of MLPs, standard MLPs struggle to capture high-frequency signals, whereas real-world signals often contain intricate and rich high-frequency components. For NeRN, it employs a 5-layer ReLU-based MLP to represent CNN kernels, balancing approximation efficiency with hardware-friendly computation. 

In this section, the output space and input coordinate transformations are optimized for Spectral Bias Suppression, with the goal of enhancing the parameter efficiency of coordinate-based ReLU-MLPs in NeRN task. The framework is shown in Fig.~\ref{fig:overflow}.

\subsection{Smoothness of CNN Kernels}
The task of representing CNN kernels can be reformulated as learning a mapping under given coordinates: given a spatial coordinate as input, the network predicts the kernel at that location. By the Weierstrass approximation theorem, any finite set of samples admits a smooth approximant, allowing us to model the spatial variation of kernels with a sufficiently smooth function. Consequently, the problem can be framed as fitting this smooth mapping with an MLP. However, a ReLU-activated MLP partitions its input space into linear regions determined by hidden-layer width; in practice, the number of regions is much smaller than the number of sample points, biasing the network toward low-frequency signals (see Eq.~\eqref{Low-Pass}). Since CNN kernels do not possess the intrinsic smoothness of natural images or videos, balancing the MLP’s approximation capacity with the kernels’ inherent structure remains a critical challenge.

Through the above analysis, it is recognized that establishing a precise definition of "kernel smoothness" is necessary, and that improving the approximation efficiency of MLPs through localized smoothness regularization is worth exploring. This problem can fundamentally be framed as bounding the approximation error incurred when fitting smooth functions within a limited number of linear regions.

Let the target function \( f: \mathbb{R}^d \to \mathbb{R}^m \) be a \( C^2 \)-smooth function, and let the sampled dataset \( \mathcal{D} = \{(\mathbf{x}_i, \mathbf{y}_i)\}_{i=1}^n \) satisfy \( \mathbf{y}_i = f(\mathbf{x}_i) \), and $ \|\mathbf{x}_{i+1} - \mathbf{x}_i\| \leq \epsilon$. Consider a ReLU-MLP \( g_\theta: \mathbb{R}^d \to \mathbb{R}^m \) with parameters \( \theta \) trained to regress \( f \), where the hidden layer width is constrained such that the number of linear regions \( K \) satisfies \( K \ll n \).

Since the MLP performs a linear approximation within each linear region, the local linearity of points in such regions is closely associated with the approximation error. To quantify this linearity, the first-order difference of the network output is applied, resulting in the following expression.
\begin{equation}
\sum_{i=1}^{n_k-1} \|\Delta\mathbf{f}_i\|_F^2 \leq \frac{m\delta^2}{\epsilon^2} \sum_{i=1}^{n_k-1} \|\mathbf{x}_{i+1} - \mathbf{x}_i\|^2 \leq m n_k \delta^2
\end{equation}
Here, the $m$ denote the output dimension determined by the CNN kernel size, $n_k$ represent the number of sampling points in region $\mathcal{R}_k$ (inversely proportional to MLP hidden unit count), and $\delta$ denotes the maximum first-order difference between adjacent samples in $\mathcal{R}_k$. 

To constrain $\delta$ in NeRN, smoothness regularization is applied to the CNN kernels. For kernels $\mathbf{w}_{l,f,c}$ (associated with layer $l$, filter $f$, and channel $c$), their smoothness is enforced through the following regularization term:
\begin{equation}
\mathcal{S}(\mathbf{w}) = \sum_{l=1}^{L-1} \sum_{f=1}^{F-1} \sum_{c=1}^{C-1} \sum_{j=1}^{3} 
\left\|\mathbf{w}_{(l,f,c)+e_j} - \mathbf{w}_{(l,f,c)}\right\|_2
\label{regularization}
\end{equation}
where $e_j$ denotes the canonical basis vector along the $j$-th spatial dimension. The proposed regularization term enforces local Lipschitz continuity of the network output by penalizing abrupt weight changes across adjacent layers, filters, and channels. In paper \cite{ref6}, smoothness is imposed via cosine similarity between neighboring CNN kernels. However, the above theoretical analysis reveals that cosine similarity fails to effectively control smoothness along individual dimensions of the kernels, resulting in increased approximation error over linear regions. In contrast, the Euclidean norm-based regularization term \(\mathcal{S}(\mathbf{w})\) provides a more precise constraint on weight variations across neighboring coordinates, leading to improved kernel smoothness and enhanced low-frequency energy.

\subsection{Unidirectional Ordering–based Smoothing}

According to Eq.~\eqref{Low-Pass}, ReLU-MLPs exhibit stronger representational capacity for low-frequency signals. However, it is observed that neural network weights are often highly non-smooth. To enable the MLP more effectively approximate such weights, enhancing their smoothness becomes crucial.

To address this issue, prior work \cite{ref6,ref36} introduced weight permutation, treating the problem as a combinatorial optimization task that searches for permutations to impose smoothness. Although conceptually straightforward, these approaches overlook a critical limitation: \textbf{interference among multi-directional optimizations leads to degradation in unidirectional smoothness metrics}.

As formalized in Eq.~\eqref{regularization}, the directional smoothness of CNN kernels along specific directions (layer, filter, channel) significantly influences the parameter efficiency of MLP. However, due to the inherent limitations of combinatorial optimization, enforcing smoothness constraints along multi-directions often results in suboptimal directional smoothness. According to Eq.~\eqref{eigendecomposition}~\eqref{Low-Pass}, optimizing smoothness along a unidirectional path introduces a spectral bias, such that:
\begin{equation}
||f_t(\mathcal{X})-\mathcal{Y}||_2 \approx |c_1|e^{-\eta\lambda_1 t} \quad (|c_1| \gg |c_2| \approx |c_3|)
\end{equation}
Here, $c_1$ denotes the projection of $\mathcal{Y}$, smoothed unidirectionally, onto the low-frequency component of the NTK, thereby reflecting its concentrated energy in the low-frequency domain. When smoothing is applied along multi-directions, although $\mathcal{Y}$ may exhibit higher energy in relatively high-frequency directions, the convergence rate is inferior to that of unidirectional smoothing due to significantly smaller associated eigenvalues $\lambda$ compared to those of low-frequency components. The dominant contribution of low-frequency components ($c_1$) under unidirectional optimization leads to faster convergence than multi-direction scenarios with more evenly distributed spectral components.

Based on the above analysis, we propose a Unidirectional Ordering–based Smoothing (UOS) method, which adopts a greedy permutation algorithm that searches for an optimal weight ordering along a unidirectional path to improve the parameter efficiency of the MLP. 

\begin{figure}[t]               
  \centering                       
  \includegraphics[width=0.9\textwidth]{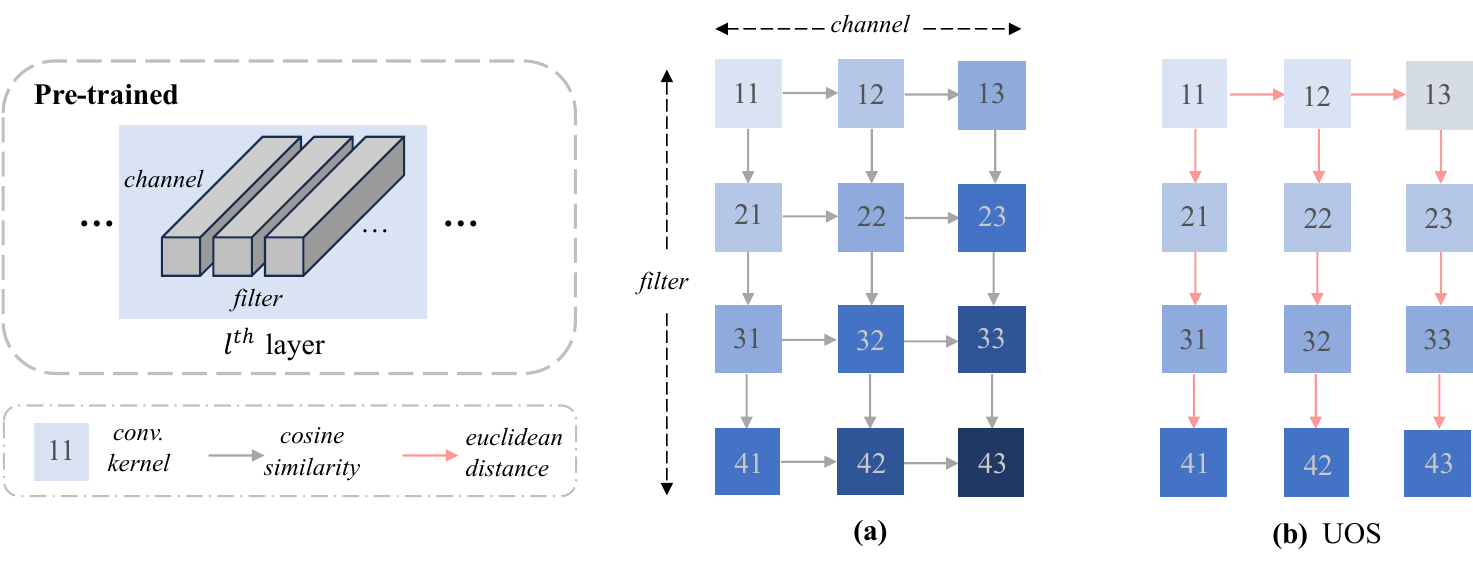}
  \caption{Different smoothing strategies applied to the CNN kernels in the $l^\text{th}$ layer of the pre-trained CNN.} 
  \label{fig:sample2}               
\end{figure}

\begin{figure}[t]               
  \centering                       
  \includegraphics[width=0.98\textwidth]{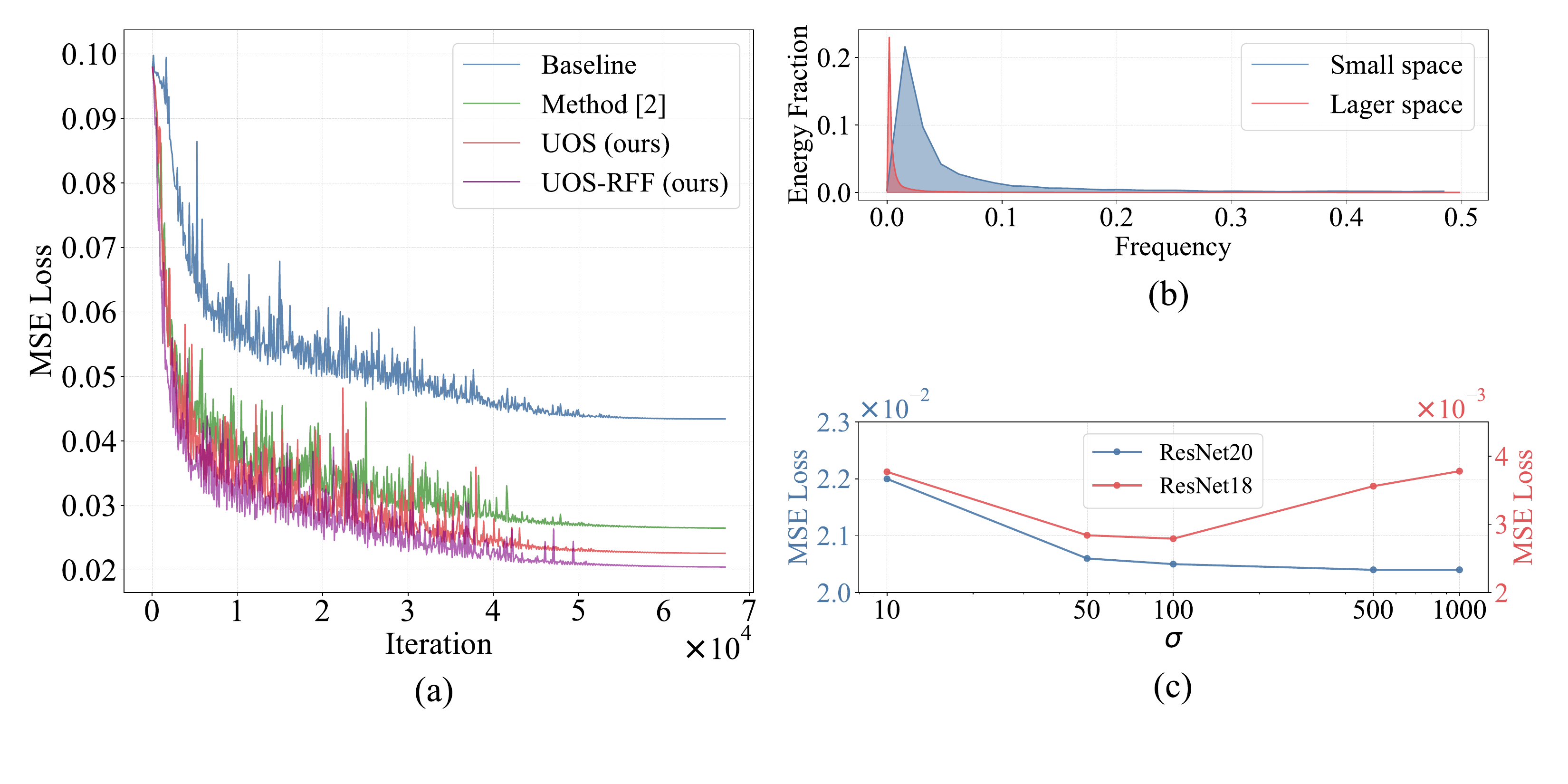}
  \vspace{-8pt}
  \caption{(a) Convergence rates of different smoothing strategies during the training of NeRN ResNet20 on CIFAR-10, with the MLP has a hidden size of 140. (b) Energy distributions under the UOS strategy for models with large and small parameter spaces. (c) Reconstruction error of NeRN ResNet18 (large parameter space) and NeRN ResNet20 (small parameter space) under UOS, using RFF with varying $\sigma$.} 
  \label{fig:sample3}               
\end{figure}

As shown in Fig.~\ref{fig:sample2}, permutation strategies for CNN kernels within a single layer are illustrated in (a), as adopted in Paper~\cite{ref6}, and (b) UOS, proposed in this work. Each kernel is indexed by its coordinate $(f, c)$, with the layer index $l$ held fixed. The approach in (a) optimizes based on the cosine similarity metric $\Delta_c$, as defined in Eq.~\eqref{eq:objective}(a), while (b) employs the euclidean distance metric, with the objective given in Eq.~\eqref{eq:objective}(b). 
\begin{equation}
\label{eq:objective}
(a) \min\sum_{f=1}^{F-1} \sum_{c=1}^{C-1}\sum_{j=1}^{2} \Delta_c\left(\mathbf{w}_{(l,f,c)+e_j},\mathbf{w}_{(l,f,c)}\right),
(b) \min\sum_{f=1}^{F-1} \left\|\,\mathbf{w}_{(l,f+1,c)} - \mathbf{w}_{(l,f,c)}\,\right\|_2 
\end{equation}
As previously discussed, our UOS, using Euclidean distance as a smoothness metric ensures per-dimension local smoothness of the CNN kernels, while unidirectional ordering facilitates achieving locally optimal smoothness. Fig.~\ref{fig:sample3}~(a) illustrates the effectiveness of our smoothing strategy, which achieves a significantly faster convergence rate than the baseline (without smoothing) and also outperforms the method proposed in NeRN \cite{ref6}, demonstrating a stronger capability in suppressing spectral bias.

\subsection{UOS-aware Random Fourier Features}
Prior work \cite{ref15} has shown that Random Fourier Features (RFF) consistently outperform fixed positional encoding 
\begin{equation}
\phi(\mathbf{x}) = [\sin(b^0\pi \mathbf{x}),\ \cos(b^0\pi \mathbf{x}),\ \ldots,\sin(b^{L-1}\pi \mathbf{x}),\ \cos(b^{L-1}\pi \mathbf{x})]
\end{equation}
across a variety of image‐regression tasks. In this paper, it is theoretically proven—and empirically confirmed—that this advantage also holds for the NeRN task.

When input coordinates are mapped via RFF, the NTK of a ReLU-MLP becomes
\begin{equation}
H(\mathbf{x_i},\mathbf{x_j}) = \frac{1}{2\pi}\phi_{ij}\bigl[\pi - \arccos(\phi_{ij})\bigr]
\end{equation}
Where \(\phi_{ij}\) is the cosine similarity of \(\phi(\mathbf{x}_i)\) and \(\phi(\mathbf{x}_j)\) under the mapping: \begin{equation}
\phi(\mathbf{x}) = [\cos(\pi\mathbf{Bx}), \sin(\pi\mathbf{Bx})], \quad \mathbf{B} \sim \mathcal{N}(0,\sigma^2\mathbf{I})
\label{RFF}
\end{equation}
As the $D \to \infty$, \(\left<\phi(\mathbf{x}_i),\phi(\mathbf{x}_j)\right>\) converges to the Gaussian kernel:
\begin{equation}
\phi_{ij} \approx \exp\left(-\frac{\sigma^2\|\mathbf{x}_i - \mathbf{x}_j\|^2}{2}\right)
\end{equation}
which preserves high-frequency sensitivity—distant points (\(\|\mathbf{x}_i - \mathbf{x}_j\|\gg0\)) still yield small \(\phi_{ij}\) enhancing locality and fine-scale discrimination. By contrast, angle-based NTKs rapidly suppress high frequencies. Crucially, the RFF bandwidth $\sigma^2$ directly tunes the kernel’s spectral decay, whereas fixed frequency grids suffer exponential sparsity and poor high-frequency coverage. Numerical experiments are conducted, as detailed in Fig.~\ref{fig:sample1}(C)–(F).

Additionally, it is shown that under UOS, network weights exhibit a clear shift toward lower-frequency components, and this bias is intensified as the parameter space size increases, as illustrated in Fig.~\ref{fig:sample3} (b). Building on this insight, We propose the UOS-aware Random Fourier Features variant, named UOS-RFF. The key design principle: for wider CNNs—where layer-wise parameter counts grow—you should reduce RFF bandwidth ($\sigma^2 \downarrow$) to rebalance frequency representation under UOS, as detailed in Fig.~\ref{fig:sample3} (c).

\section{Experiment}

In this section, we evaluate our method on three widely used image classification benchmarks: CIFAR-10, CIFAR-100 \cite{ref37}, and ImageNet \cite{ref38}. We also focus on ResNet \cite{ref39} architectures, following the similar setup in NeRN \cite{ref6}. All experiments are conducted using PyTorch on a single NVIDIA V100 GPU. The CIFAR experiments take approximately 2–6 hours, depending on the size of the original model and MLP of INR. ImageNet experiments require around 30–50 hours, varying with the hidden size of MLP.

\begin{table}[h]
\captionsetup{skip=6pt} 
\caption{CIFAR-10 reconstruction results}
\label{cifar10}
\centering
\begin{tabular}{C{2.2cm} C{2.2cm} C{2.2cm} C{2cm} C{2cm}}
\toprule
\makecell{Achitecture\\(Accuracy \%)} & \makecell{Permutation\\Smothness} & \makecell{Hidden (CR) } & \makecell{NeRN\\Accuracy \%} & Ours \\
\midrule
\multirow{6}{*}{\makecell{ResNet20\\(91.69)}}&\multirow{3}{*}{\makecell{None}}&140 (35\%)&87.95&$\textbf{89.23}$\\
&&160 (43\%)&89.90&$\textbf{90.52}$\\
&&180 (53\%)&90.82&$\textbf{91.22}$\\
\cmidrule(l){2-5}

&\multirow{3}{*}{\makecell{In-layer}}&
100 (20\%)&87.42&$\textbf{90.50}$\\
&&120 (27\%)&89.22&$\textbf{91.04}$\\
&&140 (35\%)&90.39&$\textbf{91.50}$\\

\midrule
\multirow{6}{*}{\makecell{ResNet56\\(93.52)}}&\multirow{3}{*}{\makecell{None}}&240 (27\%)&87.54&\textbf{90.27}\\
&&280 (36\%)&90.83&\textbf{91.93}\\
&&320 (45\%)&92.11&\textbf{92.73}\\

\cmidrule(l){2-5}
&\multirow{3}{*}{\makecell{In-layer}}&200 (20\%)&90.64&\textbf{92.51}\\
&&240 (27\%)&91.79&\textbf{92.96}\\
&&280 (36\%)&92.45&\textbf{93.10}\\

\bottomrule
\end{tabular}
\end{table}

\subsection{CIFAR-10}
We begin by training ResNet-20 and ResNet-56 to serve as the original backbone networks for our experiments. In Sections 4.2 and 4.3, we present UOS and RFF as structural optimizations applied to the output and input spaces, respectively, to mitigate the spectral bias of the MLP in NeRN. Table~\ref{cifar10} reports the performance gains on CIFAR-10 achieved by incorporating these techniques into the reconstruction process. "Hidden" denotes the hidden layer width of the MLP, and the compression ratio is defined as $\text{CR} = (\text{MLP Size} / \text{Model Size}) \times 100\%$. The parameter efficiency of the MLP reflects its ability to reconstruct high-quality weights at a given CR.

We evaluate reconstruction performance with and without smoothing. When permutation smoothing is “None” (only the input space is optimized), our RFF-based encoding substantially outperforms the PE used in NeRN~\cite{ref6}. Furthermore, When permutation smoothing is set to “In-layer”, it indicates that the smoothing operation is performed within each individual layer, and our proposed UOS strategy consistently delivers superior performance compared to NeRN’s original smoothing method. With only 20\% of the original model's parameters, our MLP is able to reconstruct models with an accuracy drop close to 1\%. 

\vspace{-18pt}
\begin{figure}[h]
  \centering
  \begin{minipage}[h]{0.46\textwidth}
    \centering
    \includegraphics[width=\textwidth]{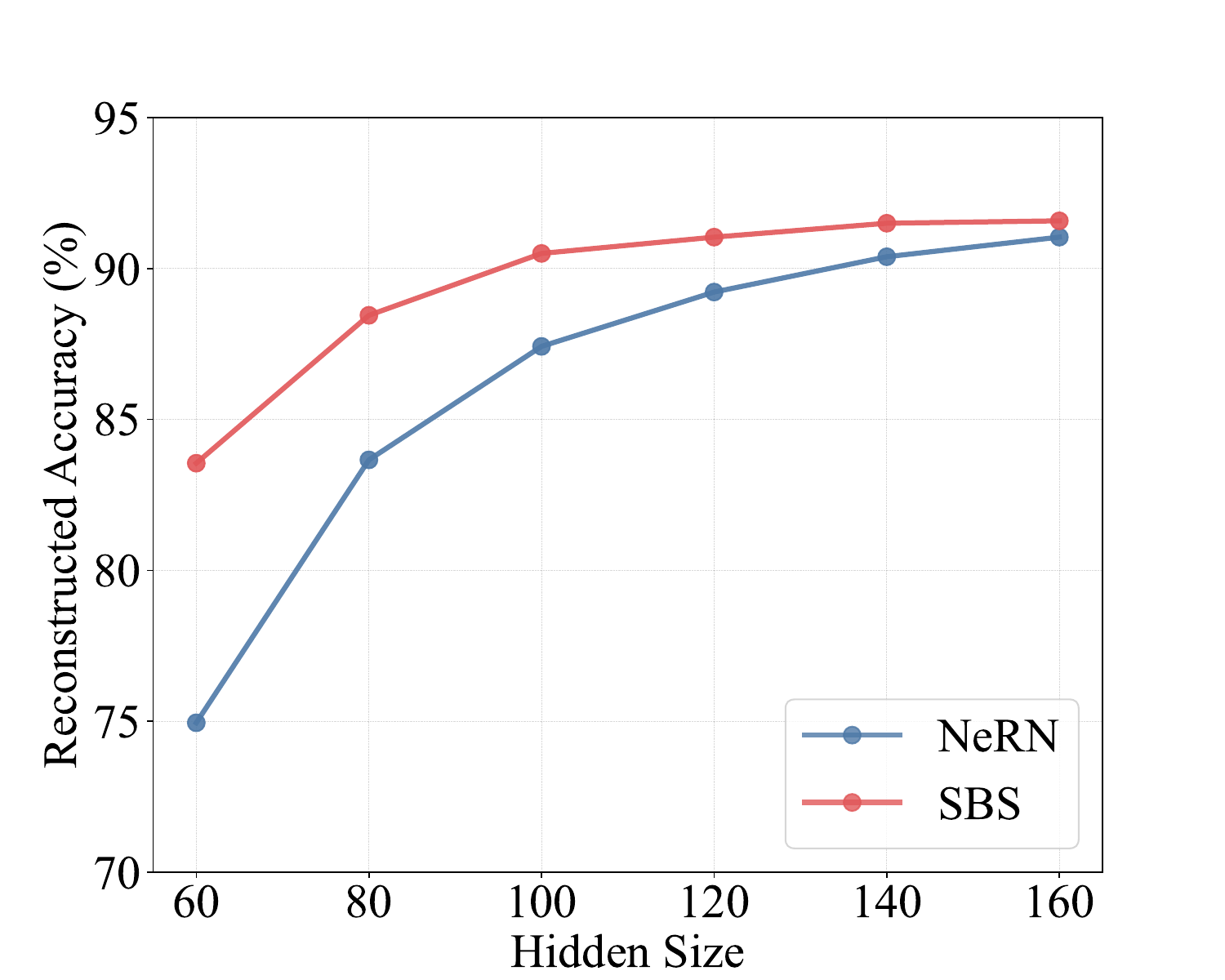}
    {\captionsetup{labelfont=bf, font=small}  
    \caption{Reconstruction performance on ResNet20 with varying MLP sizes for SBS (Ours) and NeRN.}}
    \label{fig:cifar10}
  \end{minipage}
  \hfill
  \begin{minipage}[h]{0.5\textwidth}
  \normalsize
  \vspace{-10pt}
  Fig.5 shows that under smoothing, our method achieves better parameter efficiency compared to NeRN, and our method becomes more effective when the MLP size is smaller. The bandwidth ($\sigma$) of our RFF is set to 400.

  Compared to non-smoothing, smoothing requires additional storage of a permutation index, but this index table only takes up about 4\% of the total number of parameters~\cite{ref6}, and the cost is fully controllable. 
  \end{minipage}
\end{figure}


\begin{table}[htbp]
\captionsetup{skip=6pt} 
\caption{CIFAR-100 reconstruction results}
\label{cifar100}
\centering
\begin{tabular}{C{2.2cm} C{2.2cm} C{2.2cm} C{2cm} C{2cm}}
\toprule
\makecell{Achitecture\\(Accuracy \%)} & \makecell{Permutation\\Smothness} & \makecell{Hidden (CR) } & \makecell{NeRN\\Accuracy \%} & Ours \\
\midrule
\multirow{6}{*}{\makecell{ResNet56\\(71.35)}}&\multirow{3}{*}{\makecell{None}}&320 (45\%)&68.76&\textbf{69.54}\\
&&360 (56\%)&70.09&\textbf{70.49}\\
&&400 (68\%)&70.83&\textbf{70.92}\\
\cmidrule(l){2-5}

&\multirow{3}{*}{\makecell{In-layer}}&280 (36\%)&67.06&\textbf{69.96}\\
&&320 (45\%)&69.30&\textbf{70.90}\\
&&360 (56\%)&70.31&\textbf{71.32}\\

\bottomrule
\end{tabular}
\end{table}

\subsection{CIFAR-100}
We begin by training ResNet56 as the original network, using a setup consistent with that of CIFAR-10. In ResNet56, the layer with the largest number of kernels contains the same number of kernels as the largest layer in ResNet20. Therefore, we retain the same RFF bandwidth parameter, setting $\sigma$ to 400. 
The experimental results follow a similar trend—enhancing smoothness and increasing predictor capacity lead to improved reconstruction performance, and both UOS and UOS-RFF yield substantial gains in MLP parameter efficiency over the baseline.   
As shown in Table~\ref{cifar100}, due to the greater complexity of this task compared to CIFAR-10, a slightly larger MLP is required to achieve comparable reconstruction performance.

\vspace{-20pt}
\begin{table}[h]
\centering
\captionsetup{skip=6pt}
\begin{threeparttable}
\caption{ImageNet reconstruction results}
\label{imagenet}
\begin{tabular}{C{2cm} C{2cm} C{2cm} C{2.2cm} C{2.2cm}}
\toprule
\makecell{Architecture\\(Accuracy \%)} & \makecell{Permutation\\Smoothness} & \makecell{Hidden (CR)} & \makecell{NeRN\\Accuracy \%} & Ours \\
\midrule
\multirow{5}{*}{\makecell{ResNet18\\(69.76)}}&\multirow{5}{*}{\makecell{In-layer}}&700 \ (15\%)&61.91/$66.48^*$&\textbf{68.57}\\
&&1024 (31\%)&67.55/$68.68^*$&\textbf{69.20}\\
&&1140 (38\%)&\hspace*{-7.7mm}68.21/-&\textbf{69.36}\\
&&1256 (46\%)&\hspace*{-7.7mm}68.74/-&\textbf{69.43}\\
&&1372 (55\%)&69.07/$69.32^*$&\textbf{69.43}\\
\bottomrule
\end{tabular}
\begin{tablenotes}
\footnotesize
\item[*] distillation from a stronger teacher \cite{ref8}.
\end{tablenotes}
\end{threeparttable}
\end{table}
\vspace{-20pt}

\subsection{ImageNet}
Here we show the efficiency of SBS by learning to represent the ImageNet-pretrained ResNet18. We also adopt the approach proposed in \cite{ref8} by distilling the reconstructed model using a stronger teacher network. Without any supervision, our method achieves better reconstruction performance than the version using a stronger teacher network at the same MLP size. Moreover, our approach reaches the performance of NeRN at 46\% parameter usage with only about 15\% of the MLP parameters, achieving 2×–3× improvement in parameter efficiency.

As the model size of ResNet18 is approximately 45MB and its layers are larger than those of ResNet20 (1MB), we set the bandwidth of RFF to 100. Relevant experiments are presented in Table~\ref{Ablation2}.

\subsection{Ablation Experiments}
We perform ablation studies on the UOS strategy. Table~\ref{Ablation1} reports the reconstruction performance on CIFAR-10 with ResNet20 under different smoothing strategies and input encodings, with hidden size of 140. Here, MOS (Multi-directional Ordering-based Smoothing) extends UOS by incorporating smoothness along multi-directions (filter and channel).
The data with gray background indicates the MSE between the reconstructed kernel and the original kernel. The experimental results fully verify the effectiveness of UOS.

\vspace{-12pt}
\begin{table}[t]
\centering
\captionsetup{skip=6pt}

\caption{CIFAR-10 reconstruction results with different smoothing strategies}
\label{Ablation1}
\begin{tabular}{C{2.5cm} C{2cm} C{2cm} C{2cm} C{2cm}}
\toprule
\makecell{Architecture\\(Accuracy \%)} &{\makecell{Input\\Encoding}}& \makecell{NeRN\\(MSE/Acc.\%)} & \multicolumn{1}{c}{MOS}&\multicolumn{1}{c}{UOS}\\
\midrule
\multirow{4}{*}{\makecell{ResNet20\\(91.69)}}&\multirow{2}{*}{PE}&\cellcolor{gray!15}0.0265&\cellcolor{gray!15}0.0246&\cellcolor{gray!15}0.0226\\
&&90.39&90.62&91.15\\
\cmidrule(l){2-5}
&\multirow{2}{*}{RFF}&\cellcolor{gray!15}0.0239&\cellcolor{gray!15}0.0252&\cellcolor{gray!15}0.0204\\
&&90.77&90.88&91.5\\
\bottomrule
\end{tabular}
\end{table}
\vspace{-12pt}

\newpage
Table~\ref{Ablation2} shows the impact of RFF bandwidth selection when using UOS with models of different sizes, providing more details of Fig.~\ref{fig:sample3}(c). Larger models require a smaller bandwidth to adequately capture the low-frequency components when using UOS. The MLP hidden size for ResNet20 on CIFAR-10 is 160, and for ResNet18 on ImageNet is 1024.

\vspace{-12pt}
\begin{table}[htbp]
\centering
\captionsetup{skip=6pt}
\caption{Reconstruction results with different RFF bandwidth}
\label{Ablation2}
\begin{tabular}{C{2.5cm} C{1.5cm} C{1.5cm} C{1.5cm} C{1.5cm} C{1.5cm}}
\toprule
\multirow{1}{*}{\makecell{Architecture\\(Acc.\%/Size MB)}} & \multicolumn{5}{c}{\large{$\sigma$}}\\
\cmidrule(l){2-6}
&\makecell{10}&50&100&500&1000\\
\midrule
\multirow{2}{*}{\makecell{ResNet20\\(91.69/1.04)}}&\cellcolor{gray!15}0.0220&\cellcolor{gray!15}0.0206&\cellcolor{gray!15}0.0205&\cellcolor{gray!15}0.0203&\cellcolor{gray!15}0.0204\\
&91.05&91.11&91.25&91.28&91.44\\
\midrule
\multirow{2}{*}{\makecell{ResNet18\\(69.76/44.59)}}&\cellcolor{gray!15}0.00377&\cellcolor{gray!15}0.00284&\cellcolor{gray!15}0.00279&\cellcolor{gray!15}0.00356&\cellcolor{gray!15}0.00378\\
&68.78&69.19&69.20&68.89&69.00\\
\bottomrule
\end{tabular}
\end{table}
\vspace{-12pt}

\section{Conclusion}
In this paper, we introduced SBS, a principled extension of NeRN that improves parameter efficiency by mitigating spectral bias through output smoothing (UOS) and input encoding (UOS-RFF). Our method achieves superior reconstruction quality with fewer parameters, outperforming prior work both with and without external supervision.

%
%
%
\bibliographystyle{splncs04}
\bibliography{refpaper}

\begin{thebibliography}{10}
\providecommand{\url}[1]{\texttt{#1}}
\providecommand{\urlprefix}{URL }
\providecommand{\doi}[1]{https://doi.org/#1}

\bibitem{ref13}
Arora, S., Du, S.S., Hu, W., Li, Z., Wang, R.: Fine-grained analysis of optimization and generalization for overparameterized two-layer neural networks. In: Proceedings of the 36th International Conference on Machine Learning. pp. 322--332. PMLR (2019)

\bibitem{ref6}
Ashkenazi, M., Rimon, Z., Vainshtein, R., Levi, S., Richardson, E., Mintz, P., Treister, E.: Nern: Learning neural representations for neural networks. In: The Eleventh International Conference on Learning Representations (2023)

\bibitem{ref26}
Ashkenazi, M., Treister, E.: Towards croppable implicit neural representations. In: Advances in Neural Information Processing Systems. vol.~37 (2024)

\bibitem{ref12}
Basri, R., Jacobs, D., Kasten, Y., Kritchman, S.: The convergence rate of neural networks for learned functions of different frequencies. In: Advances in Neural Information Processing Systems. vol.~32 (2019)

\bibitem{ref20}
Bemana, M., Myszkowski, K., Seidel, H.P., Ritschel, T.: X-fields: Implicit neural view-, light- and time-image interpolation. ACM Transactions on Graphics (Proc. SIGGRAPH Asia 2020)  \textbf{39}(6),  1--15 (2020). \doi{10.1145/3414685.3417827}

\bibitem{ref5}
Chen, H., He, B., Wang, H., Ren, Y., Lim, S.N., Shrivastava, A.: Nerv: Neural representations for videos. In: Advances in Neural Information Processing Systems. vol.~34, pp. 21557--21568 (2021)

\bibitem{ref16}
Chen, Y., Liu, S., Wang, X.: Learning continuous image representation with local implicit image function. In: Proceedings of the IEEE/CVF Conference on Computer Vision and Pattern Recognition. pp. 8628--8638 (2021)

\bibitem{ref23}
Chen, Z., Zhang, H.: Learning implicit fields for generative shape modeling. arXiv preprint arXiv:1812.02822  (2018)

\bibitem{ref8}
Choi, H., Thiagarajan, J.J., Glatt, R., Liu, S.: Enhancing accuracy and parameter-efficiency of neural representations for network parameterization. arXiv preprint arXiv:2407.00356  (2024)

\bibitem{ref10}
Cybenko, G.: Approximation by superpositions of a sigmoidal function. Mathematics of Control, Signals and Systems  \textbf{2}(4),  303--314 (1989)

\bibitem{ref38}
Deng, J., Dong, W., Socher, R., Li, L.J., Li, K., Fei-Fei, L.: Imagenet: A large-scale hierarchical image database. In: Proceedings of the IEEE Conference on Computer Vision and Pattern Recognition (CVPR). pp. 248--255 (2009)

\bibitem{ref28}
Deutsch, L.: Generating neural networks with neural networks. arXiv preprint arXiv:1801.01952  (2018)

\bibitem{ref4}
Dupont, E., Goli{\'n}ski, A., Alizadeh, M., Teh, Y.W., Doucet, A.: Coin: Compression with implicit neural representations. arXiv preprint arXiv:2103.03123  (2021)

\bibitem{ref27}
Ha, D., Dai, A., Le, Q.V.: Hypernetworks. arXiv preprint arXiv:1609.09106  (2016)

\bibitem{ref39}
He, K., Zhang, X., Ren, S., Sun, J.: Deep residual learning for image recognition. In: Proceedings of the IEEE Conference on Computer Vision and Pattern Recognition (CVPR). pp. 770--778 (2016)

\bibitem{ref11}
Hornik, K., Stinchcombe, M., White, H.: Multilayer feedforward networks are universal approximators. Neural Networks  \textbf{2}(5),  359--366 (1989)

\bibitem{ref2}
Jiang, C.M., Sud, A., Makadia, A., Huang, J., Nie{\ss}ner, M., Funkhouser, T.: Local implicit grid representations for 3d scenes. In: Proceedings of the IEEE/CVF Conference on Computer Vision and Pattern Recognition. pp. 6001--6010 (2020)

\bibitem{ref19}
Kim, C., Lee, J., Shin, J.: Zero-shot blind image denoising via implicit neural representations. arXiv preprint arXiv:2204.02405  (2022)

\bibitem{ref7}
Kim, S., Kim, S., Jo, W., Kim, S., Hong, S., Yoo, H.J.: C-transformer: A 2.6-18.1$\mu$j/token homogeneous dnn-transformer/spiking-transformer processor with big-little network and implicit weight generation for large language models. In: 2024 IEEE International Solid-State Circuits Conference (ISSCC). pp. 320--322. IEEE (2024)

\bibitem{ref30}
Knyazev, B., Drozdzal, M., Taylor, G.W., Romero~Soriano, A.: Parameter prediction for unseen deep architectures. In: Advances in Neural Information Processing Systems. vol.~34, pp. 29433--29448 (2021)

\bibitem{ref37}
Krizhevsky, A.: Learning multiple layers of features from tiny images. Tech. rep., University of Toronto (2009)

\bibitem{ref21}
Lin, Y.C., Florence, P., Barron, J.T., Rodriguez, A., Isola, P., Lin, T.Y.: inerf: Inverting neural radiance fields for pose estimation. In: 2021 IEEE/RSJ International Conference on Intelligent Robots and Systems (IROS). pp. 1323--1330 (2021). \doi{10.1109/IROS51168.2021.9636708}

\bibitem{ref22}
Mescheder, L., Oechsle, M., Niemeyer, M., Nowozin, S., Geiger, A.: Occupancy networks: Learning 3d reconstruction in function space. In: Proceedings of the IEEE/CVF Conference on Computer Vision and Pattern Recognition. pp. 4460--4470 (2019). \doi{10.1109/CVPR.2019.00459}

\bibitem{ref3}
Mildenhall, B., Srinivasan, P.P., Tancik, M., Barron, J.T., Ramamoorthi, R., Ng, R.: Nerf: Representing scenes as neural radiance fields for view synthesis. In: European Conference on Computer Vision. pp. 405--421 (2020)

\bibitem{ref17}
Nguyen, Q.H., Beksi, W.J.: Single image super-resolution via a dual interactive implicit neural network. In: Proceedings of the IEEE/CVF Winter Conference on Applications of Computer Vision. pp. 4936--4945 (2023)

\bibitem{ref25}
Park, J.J., Florence, P., Straub, J., Newcombe, R., Lovegrove, S.: Deepsdf: Learning continuous signed distance functions for shape representation. In: Proceedings of the IEEE/CVF Conference on Computer Vision and Pattern Recognition. pp. 165--174 (2019)

\bibitem{ref24}
Peng, S., Niemeyer, M., Mescheder, L., Pollefeys, M., Geiger, A.: Convolutional occupancy networks. In: European Conference on Computer Vision. pp. 523--540. Springer (2020)

\bibitem{ref9}
Rahaman, N., Baratin, A., Arpit, D., Draxler, F., Lin, M., Hamprecht, F.A., Bengio, Y., Courville, A.: On the spectral bias of neural networks. In: Proceedings of the 36th International Conference on Machine Learning. pp. 5301--5310. PMLR (2019)

\bibitem{ref18}
Sitzmann, V., Martel, J.N., Bergman, A., Lindell, D.B., Wetzstein, G.: Implicit neural representations with periodic activation functions. Advances in Neural Information Processing Systems  \textbf{33},  7462--7473 (2020)

\bibitem{ref1}
Sitzmann, V., Zollh{\"o}fer, M., Wetzstein, G.: Scene representation networks: Continuous 3d-structure-aware neural scene representations. In: Advances in Neural Information Processing Systems. vol.~32 (2019)

\bibitem{ref31}
Soro, B., Andreis, B., Lee, H., Chong, S., Hutter, F., Hwang, S.J.: Diffusion-based neural network weights generation. arXiv preprint arXiv:2402.18153  (2024)

\bibitem{ref15}
Tancik, M., Srinivasan, P.P., Mildenhall, B., Fridovich-Keil, S., Raghavan, N., Singhal, U., Ramamoorthi, R., Barron, J.T., Ng, R.: Fourier features let networks learn high frequency functions in low dimensional domains. Advances in Neural Information Processing Systems  \textbf{33} (2020)

\bibitem{ref36}
Yang, X., Wang, X.: Neural metamorphosis. In: Computer Vision – ECCV 2024. Lecture Notes in Computer Science, vol. 15092, pp. 1--19. Springer, Cham (2025)

\bibitem{ref29}
Zhang, C., Ren, M., Urtasun, R.: Graph hypernetworks for neural architecture search. arXiv preprint arXiv:1810.05749  (2018)

\bibitem{ref14}
Zhong, E.D., Bepler, T., Davis, J.H., Berger, B.: Reconstructing continuous distributions of 3d protein structure from cryo-em images. In: Proceedings of the International Conference on Learning Representations (ICLR) (2020)

\end{thebibliography}
\end{document}